\documentclass[letterpaper, 10 pt, conference]{ieeeconf}  % Comment this line out if you need a4paper

\IEEEoverridecommandlockouts                              % This command is only needed if 
\usepackage{cite}
\usepackage{amsmath,amsfonts}
\usepackage{algorithmic}
\usepackage{algorithm}
\usepackage{array}
\usepackage{textcomp}
\usepackage{stfloats}
\usepackage{url}
\usepackage{verbatim}
\usepackage{graphicx}
\usepackage{booktabs} 
\usepackage{svg}
\usepackage{diagbox}
\usepackage{makecell} % 在导言区导入makecell宏包
\usepackage{multirow}  % 需要加载 multirow 宏包
\usepackage{caption}
\usepackage[font=footnotesize]{subcaption}
\usepackage{booktabs}

\title{\LARGE \bf
DQO-MAP: Dual Quadrics Multi-Object mapping with Gaussian Splatting
}

\author{
    Haoyuan Li\textsuperscript{1}, 
    Ziqin Ye\textsuperscript{1}, 
    Yue Hao\textsuperscript{1}, 
    Weiyang Lin\textsuperscript{1}, 
    Chao Ye\textsuperscript{1,†}% <- 这个 % 阻止空格
    \thanks{*Corresponding author: Chao Ye, email: yechao@hit.edu.cn}% <- 添加通信作者说明
    \thanks{\textsuperscript{1}The authors are with Research Institute of Intelligent Control and Systems, Harbin Institute of Technology, Harbin 150001, China}%
}

\begin{document}

\maketitle
\thispagestyle{empty}
\pagestyle{empty}

%%%%%%%%%%%%%%%%%%%%%%%%%%%%%%%%%%%%%%%%%%%%%%%%%%%%%%%%%%%%%%%%%%%%%%%%%%%%%%%%
\begin{abstract}

Accurate object perception is essential for robotic applications such as object navigation. In this paper, we propose DQO-MAP, a novel object-SLAM system that seamlessly integrates object pose estimation and reconstruction. We employ 3D Gaussian Splatting for high-fidelity object reconstruction and leverage quadrics for precise object pose estimation. Both of them management is handled  on the CPU, while optimization is performed on the GPU, significantly improving system efficiency.  By associating objects with unique IDs, our system enables rapid object extraction from the scene.  Extensive experimental results on object reconstruction and pose estimation demonstrate that DQO-MAP achieves outstanding performance  in terms of precision, reconstruction quality, and computational efficiency. The code and dataset are available at: https://github.com/LiHaoy-ux/DQO-MAP.

\end{abstract}

%%%%%%%%%%%%%%%%%%%%%%%%%%%%%%%%%%%%%%%%%%%%%%%%%%%%%%%%%%%%%%%%%%%%%%%%%%%%%%%%
\section{INTRODUCTION}
VISION-based Simultaneous Localisation and Mapping (SLAM)  plays a crucial role in the field of AR/VR and robotics. Previous works concentrated to provide accurate ego estimation and environment maps for navigation. However, these maps  consisted of discrete points are only contain metric information, which  limits their application in more complex tasks, such as object navigatior that require scene understanding. With deep learning method, it is possible to introduce semantic information into the map. And Object-SLAM, which is aimed at constructing object-level map with semantic information, has attracted the interest of many researchers.

Unlike point-level maps, Object-SLAM systems\cite{yang2019cubeslam,nicholson2018quadricslam} utilize detections or semantic information to construct a map that includes objects’ poses and locations in the scene, which can be leveraged for downstream applications. However, a crucial challenge is how to  make objects representation more effective. Some researchers replace original objects with geometric primitives (e.g., cubes, quadrics). These compact representations serve as higher-dimensional features for navigation, including key information, such as category, size, pose, and location. Object-level maps offer advantages in long-term relocalization compared to point-level maps. However, geometric primitives lack valuable information about shape and texture features, posing a challenge for widespread application.

\begin{figure}[!ht]
        \centering
        \includegraphics[width=3.5in]{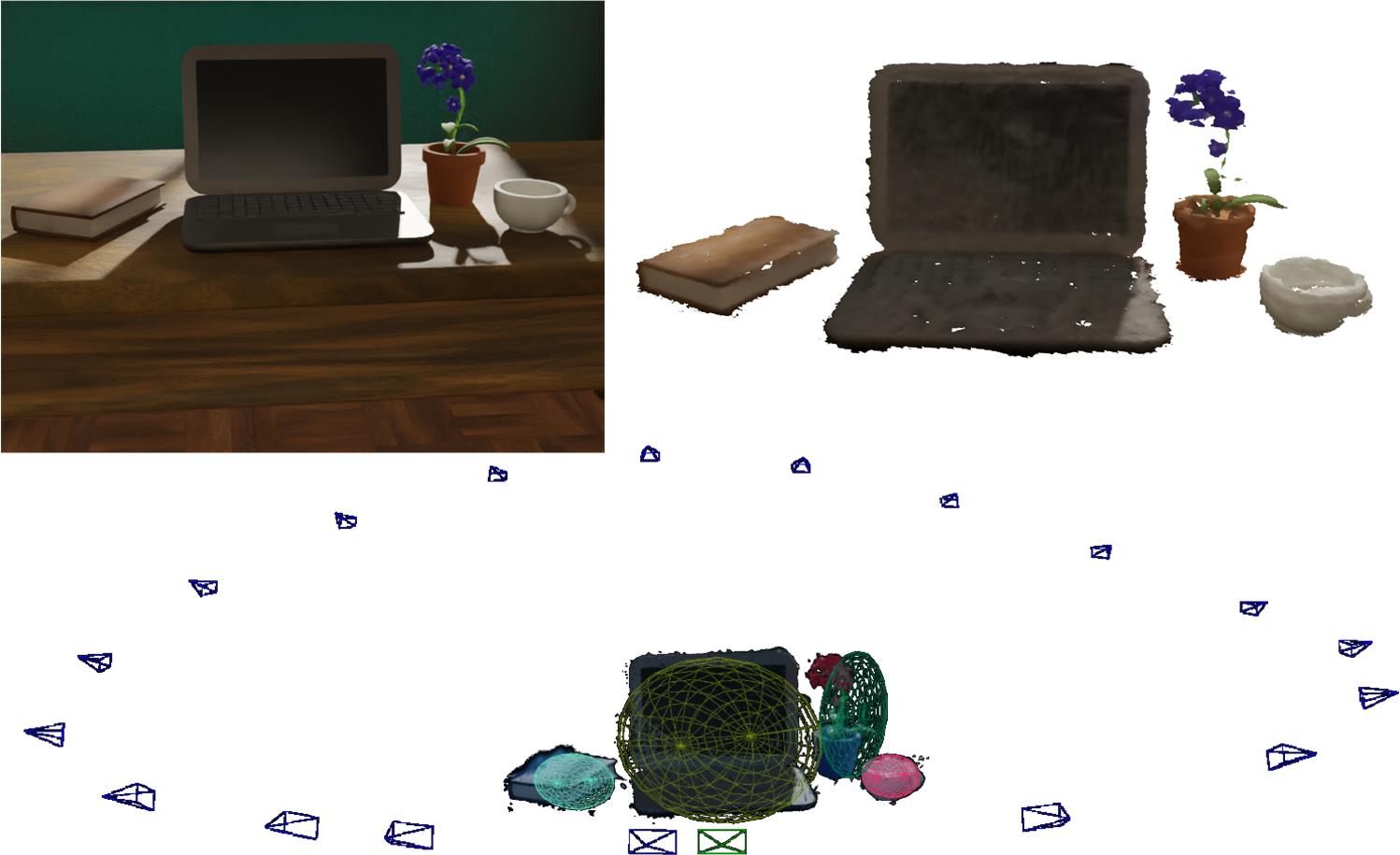}
        \caption{DQO-MAP simultaneously reconstructs objects using Gaussian Splatting and estimates their poses with quadrics.Each object is assigned a unique ID for association and extraction.}
        \label{fig6}
 \end{figure}

The reconstruction of objects' shape has been extensively studied. Some researchers have explored depth-based methods to resolve geometry, such as surfels and signed distance functions (SDFs). More recently, neural networks have gained popularity for inferring and optimizing geometry in latent spaces, which can then be decoded into voxel grids or implicit functions. While these approaches can reconstruct complete objects from sparse observations, they are often constrained by prior assumptions and struggle to handle arbitrary geometric shapes. A natural question arises: Can we reconstruct objects without prior knowledge? Neural Radiance Fields (NeRF) \cite{mildenhall2021nerf} and 3D Gaussian Splatting (3DGS) \cite{kerbl3Dgaussians} offer promising solutions, enabling geometry reconstruction from RGB and depth images alone. Compared to NeRF, 3DGS achieves higher efficiency through explicit representation and rasterization, and has recently shown excellent performance in SLAM applications.

In this paper, we present an online multi-object reconstruction system with two tightly integrated components: object pose estimation and object reconstruction.   Built on ORB-SLAM2\cite{murORB2}, the system initializes objects and Gaussians on the CPU while optimizing them in parallel on the GPU. Guided by object IDs, it enables fast object extraction from the scene.

The contributions of this work are summarized as follows:
\begin{itemize}
        \item We propose a pose-free 3D multi-object mapping system that can simultaneously perform pose estimation and reconstruction of scene objects.
        \item We present an object-level association framework that effectively aggregates different object measurements to enhance 3D-2D correspondence accuracy.
        \item  We introduce an efficient object loss function and an incremental update strategy, enabling real-time performance.
        \item Comprehensive experiments across synthetic and real-world datasets demonstrate the system's superior object perception performance.
        \end{itemize}

\section{RELATED WORK}

\subsection{Object SLAM}
As the pioneering object-based SLAM system, SLAM++ \cite{salas2013slam++} utilized depth data to match object models and optimized object maps through pose-graph optimization. However, the dependence on prior object knowledge limits its real applications. Subsequently,  researcher focus on  online object reconstruction. Fusion++ \cite{mccormac2018fusion++}, MaskFusion \cite{runz2018maskfusion} employed Mask R-CNN for instance segmentation and fused multi-frame observations to estimate TSDF, but its implementation is dependent on computational resources.

In contrast to dense reconstruction approaches, some studies explored lightweight object maps using geometric primitives. Cube SLAM \cite{yang2019cubeslam} reconstructed 3D cuboids from 2D bounding boxes (BBox), assuming coplanar object placement to simplify the optimization. Quadric SLAM \cite{nicholson2018quadricslam} parameterized objects as differentiable quadrics and optimized them via reprojection errors, but it requires manual association. To address this issue, VOOM \cite{wang2024voom} established data association relationships using ORB feature and quadric distance metrics.

Volume rendering and Gaussian splatting provide alternative methods  for  localization and reconstruction. vMAP \cite{kong2023vmap} encoded objects via MLPs, prioritizing rendering fidelity over geometric precision. RO-MAP \cite{han2023ro} decoupled construction of object maps into reconstruction with NeRF and object pose estimation used cuboid. However,  it is inherited NeRF's computational limitations and it is still a loosely coupled system. Our work tightly integrates object pose estimation and reconstruction by combining quadric parameterization with 3DGS for efficient simultaneous optimization.

\subsection{Gaussian Splatting in SLAM}

The 3DGS\cite{kerbl3Dgaussians}, as an explicit radiance field representation, outperforms NeRF in rendering speed and geometric interpretability, consequently gaining widespread adoption. However, 
3DGS still faces challenges to use directly for Object SLAM. Existing 3DGS frameworks prioritize global scene reconstruction over object-level modeling. SiLVR \cite{tao2024silvr} optimized scenes via submap stitching without isolating dynamic objects, while Mip-Splatting \cite{yu2024mip} enhanced rendering quality but lacks instance-level editing. Although these methods reconstruct entire scenes, extracting specific objects requires additional post-processing.

Another obstacle to the further utilization of 3DGS are resources and efficiency. As explicit expressions, numerous variables need to be stored and optimized. For instance, MonoGS\cite{matsuki2024gaussian} and Splatam\cite{keetha2024splatam} added and optimized Gaussian at each frame, which incurs significant overhead  in the scene. To overcome the problem, RTG-SLAM\cite{peng2024rtg} classifie and managed Gausssians into different categories, which increased speed and reduced memory consumption.

Our object-centric 3DGS-SLAM framework decouples scene and object Gaussians based on object associations, enabling parallel instance-level reconstruction.  By limiting Gaussian types and refining the update strategy, our approach retains 3DGS’s real-time rendering benefits while enhancing scene robustness through object-geometric constraints.

\section{METHOD}
\begin{figure*}[!ht]
        \centering
        \includegraphics[width=0.99\textwidth]{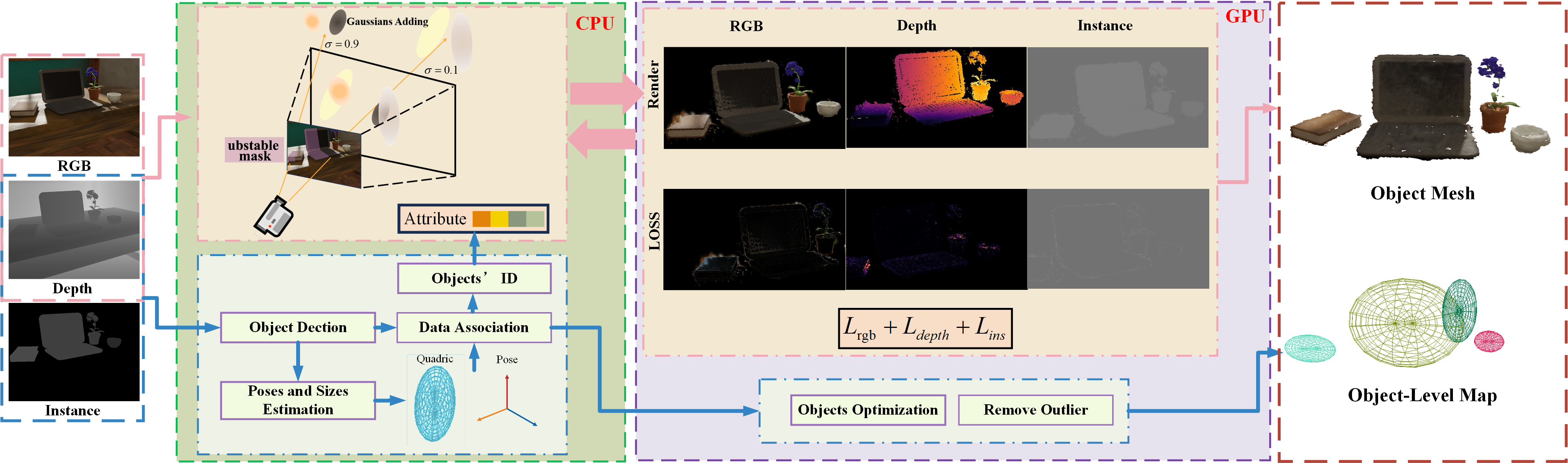}
        \caption{Overview of our proposed system. DQO-MAP tightly integrate object pose estimation and reconstruction, leveraging quadrics for object estimation and 3DGS for reconstruction.  }
        \label{flow_chart}
 \end{figure*}

The proposed method’s architecture is illustrated in Fig.\ref{flow_chart}. Our system combines object reconstruction with object-level mapping.  Given RGB-D and instance frames, it performs simultaneously object pose estimation and  reconstruction. Guided by the association results, the system enables decouple objects from the scene. Subsequently, we employ Tsdf-Fusion\cite{wei2024gsfusion} to generate a 3D mesh of the object.

 \subsection{Objects’ Pose and Size Estimation }
Unlike previous methods \cite{han2023ro,wang2024voom} that estimate only the yaw angle, our approach estimates full 6-DOF object poses. While prior work\cite{wang2023qiso}  showed that instance segmentation improves orientation accuracy, it is computationally expensive and often unstable (e.g., when using OpenCV), complicating data association. We employ YOLOv10\cite{wang2025yolov10} as the detector to generate 2D bounding boxes (BBoxes).

We use dual quadrics as a geometric primitive to estimate an object's pose. For the kth object $O_k$, it's quadric $Q_k^*$ could be calculate as:
\begin{equation} \label{eq1}
  \begin{aligned}
    \pi^T  Q_k^*\pi=0
  \end{aligned}
\end{equation}
where $\pi$ is formed by the back projection of any edge of 2D BBox to the world coordinates. As the shape of  dual quadrics is the same as 3D Gaussian, the $Q_k^*$  could be simply decoupled as follows:
\begin{equation} \label{eq2}
\begin{aligned}
\mu & = \begin{pmatrix}
  x_k & y_k & z_k
\end{pmatrix}^T \\
\Sigma&^{-1} = R(\theta)^T S^T S R(\theta)
\end{aligned}
\end{equation}
where $R(\theta)$, $\mu$ represents the rotation and location of object, respectively. And $S$ is formed by the scale $[a,b,c]$, which can be expressed as:where $R(\theta)$, $\mu$ represents the rotation and location of object, respectively. And $S$ is formed by the scale $[a,b,c]$, which can be expressed as:
\begin{equation} \label{eq3}
S=\begin{pmatrix}
  \frac{1}{a^2}&0  &0 \\
  0&\frac{1}{b^2}  &0 \\
  0&  0&\frac{1}{c^2}
\end{pmatrix}
\end{equation}

\subsection{Object-Level Association}
\begin{figure}[!ht]
        \centering
        \includegraphics[width=3.6in]{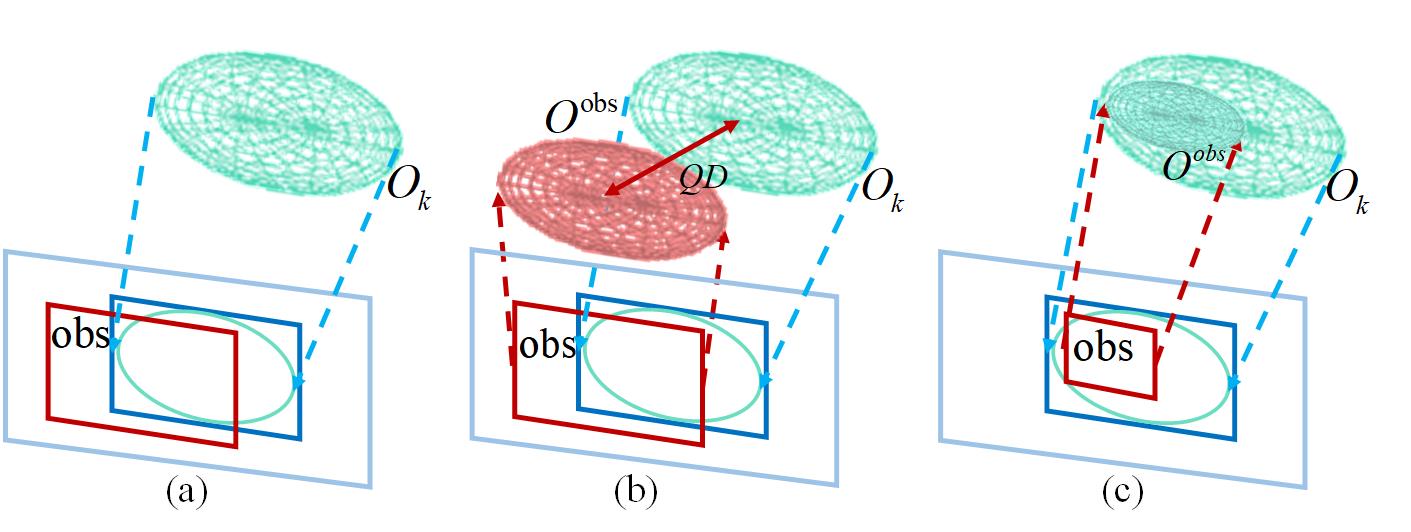}
        \caption{Different dtypes of data association}
        \label{accociation_strategy}
 \end{figure}

The object association is a key problem for the multi-object system. Unlike only using object id known in advance, we design a coarse-to-fine association strategy for 3D-2D object information association, as shown in Fig.\ref{accociation_strategy}. We first project all 3D objects to the image as follows:

\begin{equation} \label{eq4}
C_k^* = P^TQ_k^*P
\end{equation}
where $P = K \cdot Rt \in \mathbb{R} ^{3\times 4}$. 

Then, we use the Intersection over Union (IoU) between objects' projection and 2D BBoxes of each frame(in Fig.\ref{accociation_strategy}(a)) to filter the coarse results. 

All observed BBoxes matched are used for initialization object $O^{obs}$, then we use a standard distance between quadrics (QD) to filter out wrong matched(in Fig.\ref{accociation_strategy}(b)). The QD is defined as :
\begin{equation} \label{eq5}
  QD = exp(-\tau \cdot (||u^{obs} - u{k}||_2 + ||S^{obs}-S||_F))
\end{equation}
where $\tau$ is a constant and $||\cdot||_F$ is the Frobenius norm. If the $QD > thre$ ($thre$ is a threshold), the result will be filtered out.

For occlusion cases(in Fig.\ref{accociation_strategy}(c)), an object will be observed into pieces. The object's parameters will be updated to estimate the whole object.
When there are  two objects has a same id and $Area( BBox_{j}) <Area( BBox_{i})$, the score $t$ is calculated as follos:
\begin{equation} \label{eq6}
  t= OverLap(BBox_{i}, BBox_{j})/BBox_{j}
\end{equation}
If $t>t_{thre}(t_{thre}=0.85)$and $QD<d (d=0.1)$, the two objects are from the whole object, update the parameters of quadrics by pop the $O_j$.

\subsection{Gaussains Incremental Update}
Training all Gaussians per frame is inefficient and does not always enhance geometric stability. Following prior work\cite{peng2024rtg}, we adopt an incremental update strategy for classifying Gaussians as opaque  Gaussians (OG) for geometry fitting and transparent  Gaussians (TG) for color correction. Instead of updating all Gaussians, we optimize only those linked to unstable masks, determined by color, depth, and instance errors.

\begin{equation} \label{eq11}
\begin{aligned}
  Mask_{geo} &= \left\{ u_{geo} \mid \text{ins} < \theta_{\alpha}, \quad \text{or} \quad |d - \hat{d}| > \theta_d \right\}\\
  Mask_{rgb}&=\{u_c\mid |c-\hat{c}|>\theta_c \}
\end{aligned}
\end{equation}
where $Mask_{geo}$ indicates the object should be added new geometry. $ins < \theta_{\alpha}$ means the area may not remain an object, and $|d - \hat{d}| > \theta_d$ means that the object is not reconstructed completely, and $\theta_d = 0.1$.  $Mask_{rgb}$ represents wrong area, and $\theta_c = 0.1$.

 With objects' 3D-2D association across frames, we assign object IDs to every Gaussian, which is a simple but efficient way to separate objects from the scene. Through this way, the Gaussians belonging to $O_K$ need to be added or optimized could be written as:
 \begin{equation} \label{eq13}
Mask^{O_k}=\left\{    Mask_{geo}^{id} , Mask_{rgb}^{id} \mid id=k\right\}
\end{equation}

 \subsection{Object Pose and Reconstruction Training Loss}
 Object pose and Gaussians are optimized trained parallel on GPUs.
 \subsubsection{Object Pose Traing Loss} the parameters of quadrics will be optimized by minimizing the IOU loss. 
\begin{equation} \label{eq12}
  J= \sum_{i\in F_{O_k} } IOU(BBox(C_k), BBox^{obs}_i)
\end{equation}
where $F_{O_k} $ is a set of frame observe object $O_k$.
 \subsubsection{ Reconstruction Training Loss:} For a set of Gaussian, $G^k$ belonging to object $O^k$, the RGB loss is written as: 
 \begin{equation} \label{eq7}
  L_{rgb}^{obj}=\sum_{i\in G^k} ||c_i-\hat{c_i}||_2
\end{equation}
For geometry, we use a depth a loss:
\begin{equation} \label{eq8}
  L_{depth}^{obj}=\sum_{i\in G^k} ||d_i-\hat{d_i}||_2
\end{equation}
As Gaussian splatting uses $\alpha$ blending to render RGB and depth, the edges between objects and the background are blurred. We use instance information $ins \in [0,1]$ to limit the outline of the object. The $ins_i$ is calculated as:
\begin{equation} \label{eq15}
 ins = \sum_{i=1}^N \alpha_i, \alpha_i \in OG
\end{equation}
where $\alpha_i$ means Gaussain's opacity rendered on the pixel. And the instance loss $L_{ins}$ is defined as follows:
\begin{equation} \label{eq9}
  L_{ins}^{obj}=\sum ||ins_i-\hat{ins_i}||_2
\end{equation}
The overall training losses are accumulated for object $O_K$:  
\begin{equation} \label{eq10}
  L_{total}^{O_k}=L_{rgb}+L_{depth}+\lambda L_{ins}
\end{equation}
where $\lambda$ is a constant value.

%实验部分
\section{EXPERIMENTS}
We evaluate the performance of the proposed system on both synthetic and real-world datasets. We first evaluate the quality of the objects' reconstruction and then test the objects' pose. Finally, an ablation study of the state of the data association is performed.  

\textit{Implementation Detail:} The propsed system in implemented on a desktop with a  2.10Hz Intel(R) 5218R CPU, and a NVIDIA 3090 24GB GPU. We set $OG=0.9$,  $TG = 0.1$,  and the Tsdf-Fusion is used to  extract mesh, with the   $scale=0.8$.
We implement tracking, mapping and joint optimization parts with Pytorch framework, and  leverage CUDA kernels
for rasterization and back propagation. And we  used Azure Kinect RGBD camera for real-world test.

\textit{Baselines:} For reconstruction, we compare to the classical offline method,  COLMAP\cite{fisher2021colmap}, Tsdf-Fusion\cite{wei2024gsfusion}, and Nerf-based object reconstruction method, vMAP\cite{kong2023vmap}, RO-MAP\cite{han2023ro}. Addtionally, we also compoared our method with a classical GS-based approach, MonoGS\cite{matsuki2024gaussian}.
For pose estimation, we compared to RGBD object-SLAM,  VOOM\cite{wang2024voom}.

\textit{Datasets and Matrics:} For reconstruction evaluation, we evaluate the \textit{Cube-Diorama} dataset and the \textit{Replica} dataset, which provide ground truth (GT) instance segment and mesh. Accuracy, Completion are used for quantitative evaluation of object reconstruction. Subsequently, we qualitatively evaluate the system on a self-collected real-world dataset.
For object pose, we use IoU and distance to evaluate on our own simulation dataset, which is collected by \textit{Ai2-THOR}\cite{kolve2017ai2} with GT.

\subsection{Evaluation of Online Object Reconstruction}

\subsubsection{Cube-Diorama}

\begin{figure}[!ht]
        \centering
        \includegraphics[width=2.85in]{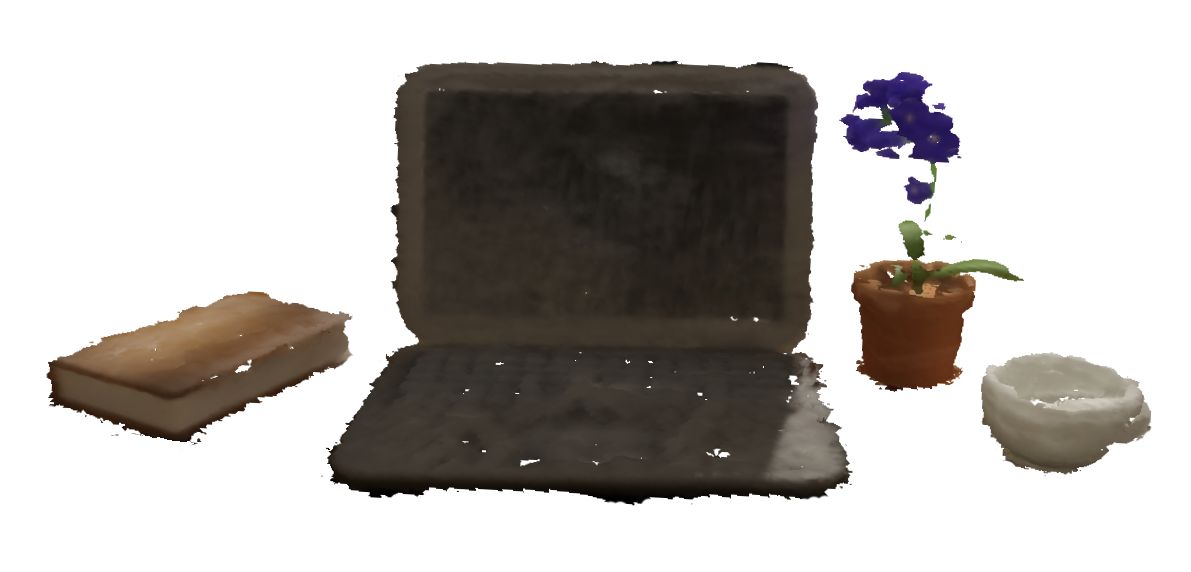}
        \includegraphics[width=2.85in]{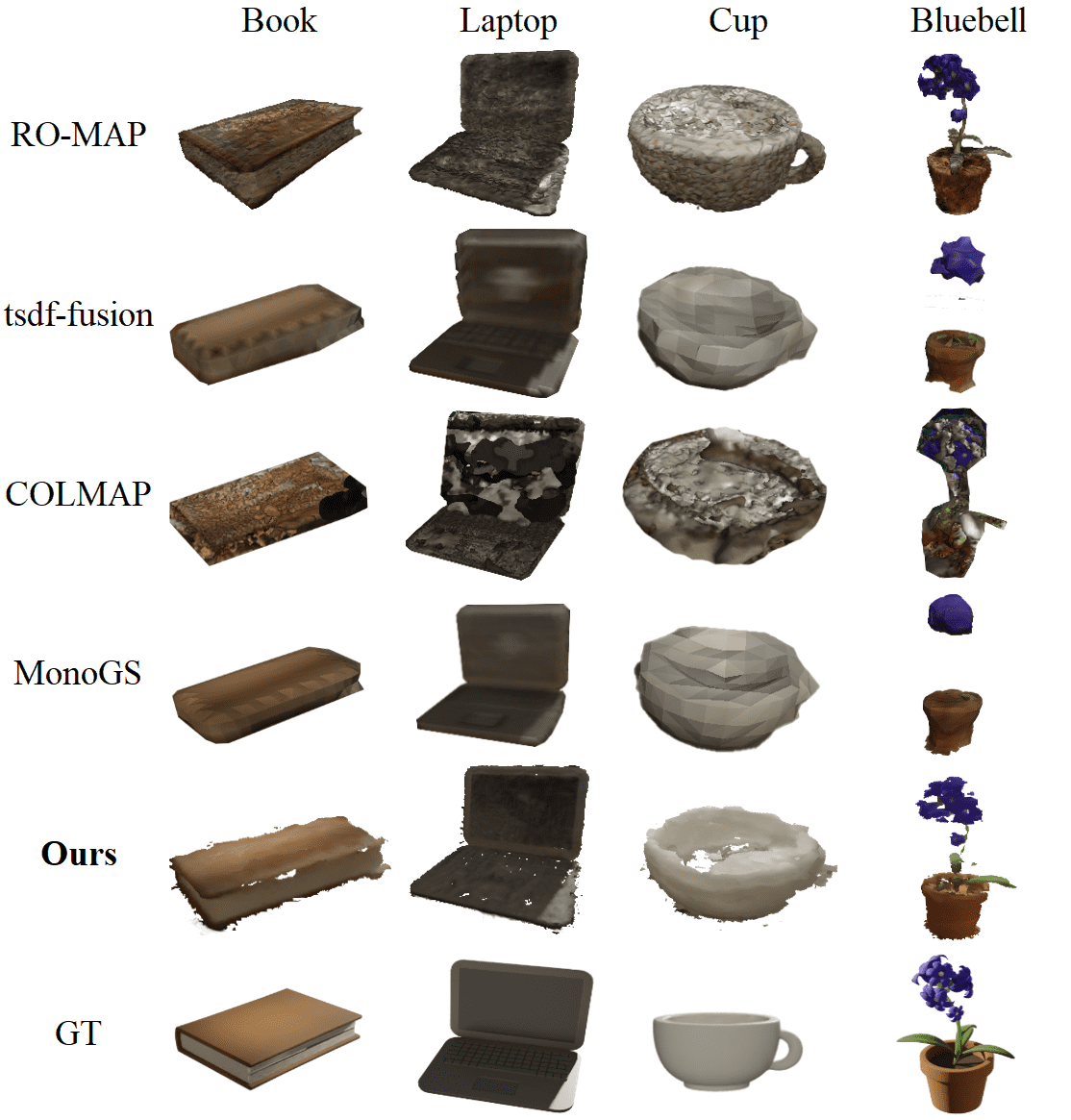}
        \caption{Object reconstruction results in \textit{Room} }
        \label{objects_Room}
 \end{figure}

 As other methods are focus on scene reconstruction, we  use post-processing to extract objects from the scene. The results,  presented in Fig.\ref{objects_Room}, indicate that the  TSDF-Fusion and MonoGS struggle with detailed texture preservation, while  RO-MAP exhibits insufficient surface smoothness and partial surface incompleteness. In contrast, our method improves boundary clarity and overall integrity while maintaining fusion-level reconstruction quality. To quantify performance, we analyzed four objects in the Room scene with Accuracy (Acc.), Completion (comp), and the results are shown in Table \ref{result_room}. Our method achieves high completeness, especially for errors under 1 cm, surpassing others in detail preservation. While RO-MAP, as a NeRF-based approach, excels in completeness, our method better preserves fine details. Overall, our results outperform other object reconstruction methods.

\begin{table}[!ht]
        \centering
        \caption{Quantitative Evaluation of Object Reconstruction on \textit{ROOM}. \textbf{Bold} and \underline{underline} indicate the best and the second-best respectively.}
        \label{result_room}
        \begin{tabular}{cccc}%p{1pt}
          \toprule
          \specialrule{0em}{3pt}{3pt}
          Method                            & Acc. [cm]↓
                       &  Comp. [cm]↓ 
                       
                       & \begin{tabular}[c]{@{}c@{}}Comp.Ratio\\ {[}\textless{}1cm\%{]↑}\end{tabular} \\
          \specialrule{0em}{3pt}{3pt}
          \hline
          \specialrule{0em}{1pt}{1pt}
          COLMAP        &3.10    & \underline{0.36}        & 91.48                         \\
          \specialrule{0em}{1pt}{1pt}
         Tsdf-Fusion     & 3.12   & 0.56   & 89.95               \\
         \specialrule{0em}{1pt}{1pt}
        RO-MAP   &  \underline{2.23}     & 0.42         & \textbf{94.34}              \\
          \specialrule{0em}{1pt}{1pt}
          MonoGS        & 3.69       & 1.04         & 66.72                     \\
          \specialrule{0em}{1pt}{1pt}
         \textbf{Ours}    &\textbf{1.92}     &\textbf{0.34}     &\underline{92.54}      \\
          \specialrule{0em}{1pt}{1pt}
          \bottomrule
        \end{tabular}
      \end{table}

\subsubsection{Replica} 
\begin{figure}[!ht]
        \centering
        \includegraphics[width=2.85in]{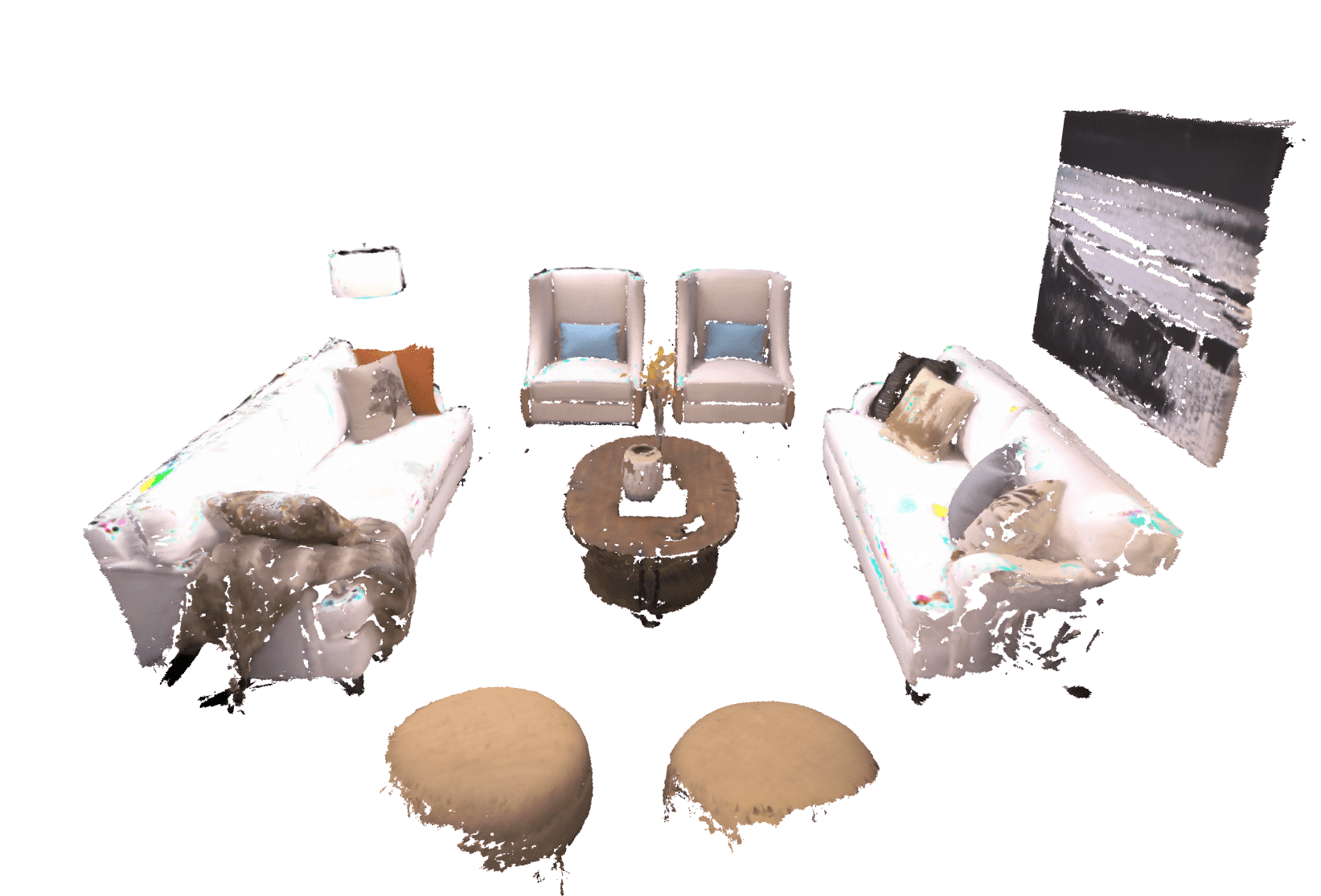}
        \includegraphics[width=2.85in]{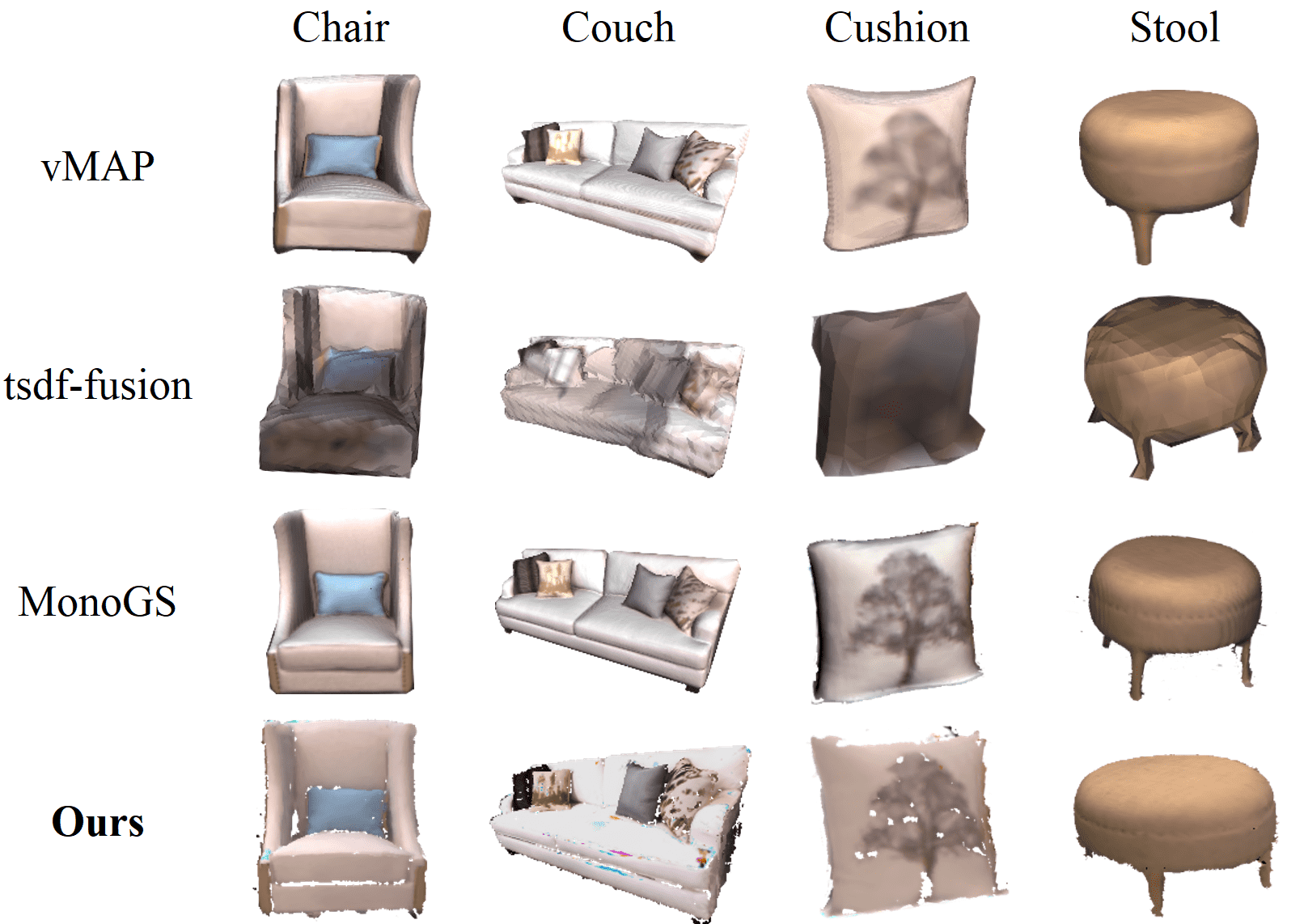}
        \caption{Object reconstruction results in \textit{Replica} }
        \label{objects_Room0}
 \end{figure}

In order to further test the performance of the proposed method in multiple objects and large scenes, we tested our method on \textit{Replica}, with results shown in Fig. \ref{objects_Room0}. TSDF-Fusion struggles with detail, texture, and structure, while vMAP and MonoGS achieve higher structural accuracy. Our method delivers superior geometric accuracy and texture fidelity while operating efficiently. In Table \ref{result_replica}, our method achieves high accuracy in indoor environments, with a notably high reconstruction rate for errors \textless 5 cm, significantly outperforming others across metrics.

\begin{table}[!ht]
\centering
\caption{Quantitative Evaluation of Object Reconstruction on \textit{Replica}.}
\label{result_replica}
\begin{tabular}{ccccc}
\toprule
Scene & Method & Acc. [cm]↓ & Comp. [cm]↓ & \begin{tabular}[c]{@{}c@{}}Comp.Ratio\\ {[\textless{}5cm\%]↑}\end{tabular} \\
\midrule
\multirow{4}{*}{\textit{room0}} & vMAP & 5.27 & 4.08 & 36.74 \\
                                & Tsdf-Fusion & 16.03 & 3.89 & 86.46 \\
                                & MonoGS & \textbf{0.62} & \textbf{0.92} & \underline{89.54} \\
                                & \textbf{Ours} & \underline{1.21} & \underline{1.07} & \textbf{90.42} \\
\midrule
\multirow{4}{*}{\textit{office0}} & vMAP & 3.15 & 15.55 & 49.09 \\
                                  & Tsdf-Fusion & 8.91 & 10.01 & \underline{82.57} \\
                                  & MonoGS & \textbf{0.61} & \underline{8.64} & 82.02 \\
                                  & \textbf{Ours} & \underline{0.82} & \textbf{4.43} & \textbf{86.21} \\
\bottomrule
\end{tabular}
\end{table}

\subsubsection{Real World}
\begin{figure}[!ht]
        \centering
        \includegraphics[width=3.6in]{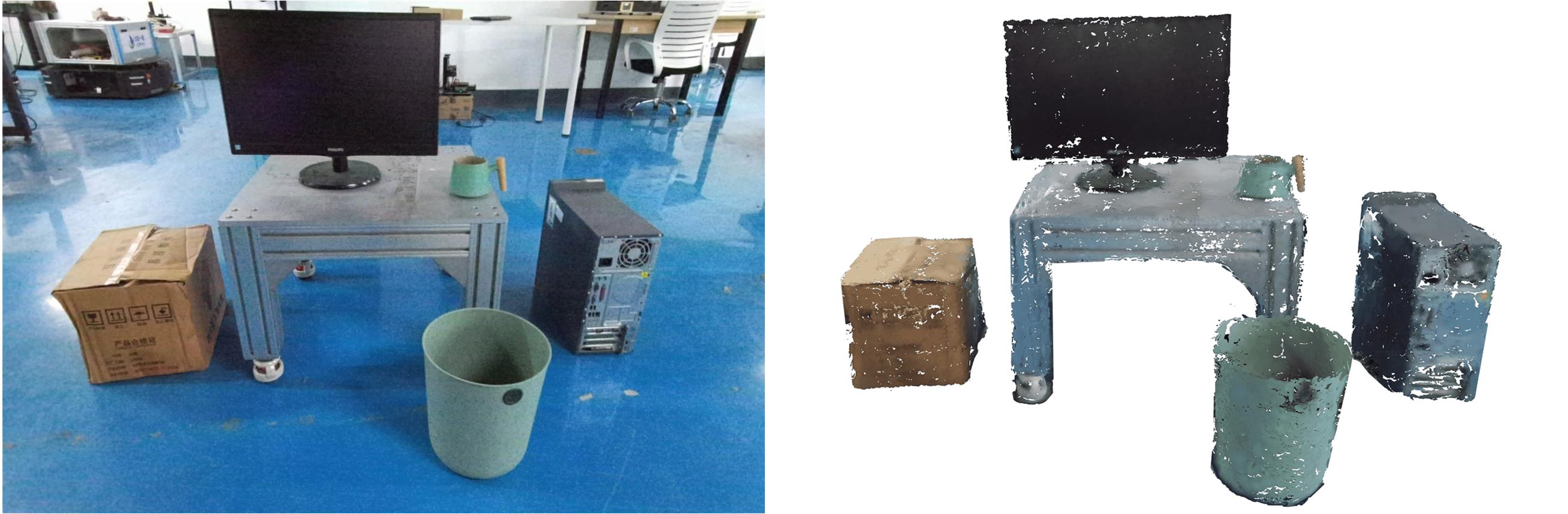}
        \caption{Object reconstruction results in real dataset}
        \label{Real}
 \end{figure}

To evaluate the robustness of our method, we conducted experiments on a real-world scene captured using the Azure Kinect RGBD camera, as shown in Fig. \ref{Real}. Given the restricted camera viewpoints and the presence of irregular objects, we conducted a qualitative analysis on the scene and single object. The results demonstrate that our method successfully extracted detailed object meshes, even in conditions with sparse data.

% \begin{figure}[!h]
%   \centering
%   \subfigure[]{
%     \includegraphics[width=4cm]{fig/real_dataset/rgb.jpg}
%   }
%   \subfigure[]{
%     \includegraphics[width=4cm]{fig/real_dataset/real_mesh.png}
%   }
%   \subfigure[]{
%     \includegraphics[width=1.2cm]{fig/real_dataset/object/box.png}
%   }
%   \subfigure[]{
%     \includegraphics[width=1.2cm]{fig/real_dataset/object/display.png}
%   }
%   \subfigure[]{
%     \includegraphics[width=1.2cm]{fig/real_dataset/object/cup.png}
%   }
%   \subfigure[]{
%     \includegraphics[width=1.2cm]{fig/real_dataset/object/pail.png}
%   }
%   \subfigure[]{
%     \includegraphics[width=1.2cm]{fig/real_dataset/object/chassis.png}
%   }
%   \caption{Object reconstruction results in real dataset}
%   \label{Real}
% \end{figure}

\subsubsection{Runtime and Memory}
In addition to reconstruction quality, we compared the average runtime and memory costs of our method with other online methods, as detailed in Table \ref{runtime}. Our method achieves faster tracking and mapping than NeRF and 3DGS methods. By incremental update strategy, we accelerated the runtime while maintaining quality.
\begin{table}[!ht]
        \centering
        \caption{Runtime and Memory Cost of  Object Reconstruction on \textit{ROOM}}
        \label{runtime}
        \begin{tabular}{ccccc}
          \toprule
          \specialrule{0em}{3pt}{3pt}
          Method                            & Tracking(s)
                       & Mapping(s)
                      & FPS(fps)             &Model Size (MB)                         \\
          \specialrule{0em}{3pt}{3pt}
          \hline
          \specialrule{0em}{1pt}{1pt}
         %vMAP         &0            &0            & 0            & 0                       \\
          \specialrule{0em}{1pt}{1pt}
         RO-MAP & \textbf{0.142}    &\underline{0.143}       & 6.99      &\textbf{806.09}      \\
         \specialrule{0em}{1pt}{1pt}
          MonoGS       & 1.2365     &  0.7773      &0.49
            &2498.52                \\
          \specialrule{0em}{1pt}{1pt}
         \textbf{Ours}                        &\underline{0.152}&\textbf{0.074}   &\textbf{13.46}& \underline{907}          \\
          \specialrule{0em}{1pt}{1pt}
          \bottomrule
        \end{tabular}
      \end{table}

\subsection{Evaluation of Object Pose Estimation}
\begin{figure}[!ht]
        \centering
        \includegraphics[width=3.3in]{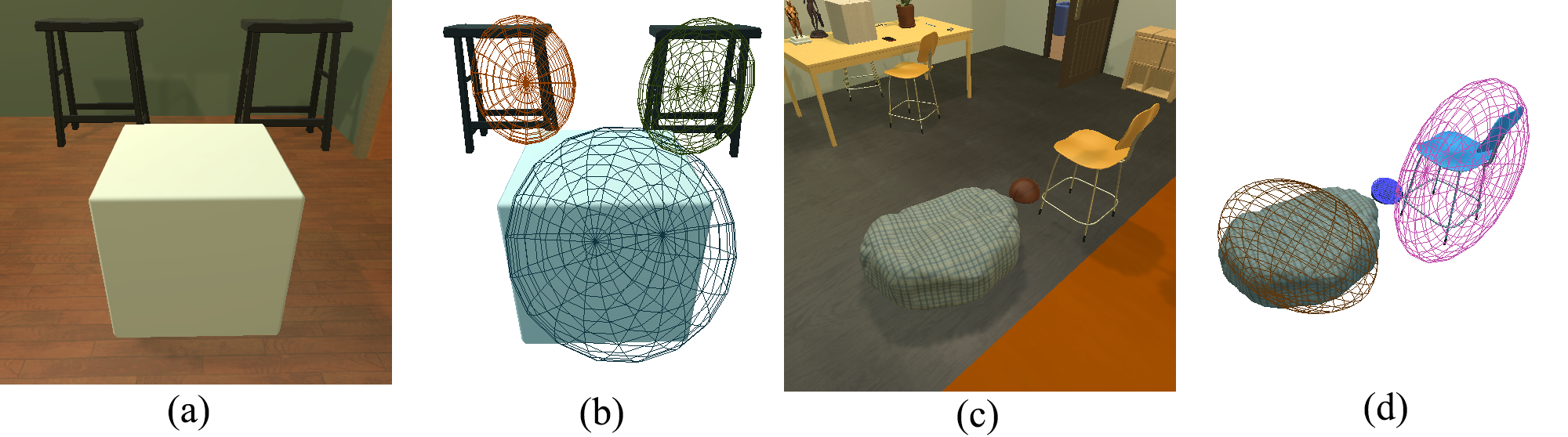}
        \caption{Evaluation of object pose in  self-collected \textit{Ai2-THOR} }
        \label{Aithor}
 \end{figure}

We evaluate the pose estimation with RGBD quadrics SLAM on synthetic dataset colllected by \textit{Ai2-THOR}\cite{kolve2017ai2} as shown in Fig.\ref{Aithor}, which provided  GT of objects. We use  Center Distance Error (CDE, cm) and IOU (2D and 3D IOU) between estimated and GT to evaluate the accuracy, and the results are shown in  Table \ref{pose_estimation}. The experimental results show that our method achieves competitive accuracy among the scene. 
\begin{table}[!ht]
        \centering
        \caption{ACCURACY OF OBJECT POSE ESTIMATION}
        \label{pose_estimation}
        \begin{tabular}{cccc}
          \toprule
          \specialrule{0em}{3pt}{3pt}
          Scene    & Metrics & VOOM& \textbf{Ours}                                 \\
          \specialrule{0em}{3pt}{3pt}
          \hline
          \specialrule{0em}{1pt}{1pt}
          \multirow{3}{*}{\textit{ROOM}}    &3D IoU↑      &0.561&\textbf{0.572}                  \\
          \specialrule{0em}{1pt}{1pt}
          &2D IoU↑         &0.650&\textbf{0.752}\\
          \specialrule{0em}{1pt}{1pt}
          &CDE↓         &0.93&\textbf{0.90}         \\
          \hline
          \specialrule{0em}{1pt}{1pt}
          \multirow{3}{*}{\textit{Ai2-THOR1}} &3D IoU↑         &0.304&\textbf{0.467}            \\
          \specialrule{0em}{1pt}{1pt}
           & 2D IoU       &0.722&\textbf{0.791}          \\
          \specialrule{0em}{1pt}{1pt}
              & CDT↓ &2.5&\textbf{1.1} \\
          \specialrule{0em}{1pt}{1pt}
          \hline
          \specialrule{0em}{1pt}{1pt}
          \multirow{3}{*}{\textit{Ai2-THOR2}} &3D IoU↑         &0.467&\textbf{0.510}             \\
          \specialrule{0em}{1pt}{1pt}
           & 2D IoU↑       &0.667&\textbf{0.723}         \\
          \specialrule{0em}{1pt}{1pt}
              & CDE↓ &1.4&\textbf{1.3} \\
          \specialrule{0em}{1pt}{1pt}
          \bottomrule
        \end{tabular}
 \end{table}

\subsection{Ablation Study}
\begin{table}[!ht]
        \centering
        \caption{DATA ASSOCIATION RESULTS}
        \label{ASSOCIATION}
        \begin{tabular}{ccccc}
          \toprule
          \specialrule{0em}{3pt}{3pt}
          Scene                          & Only IoU
                       & Only QD
                       &QD+IoU&GT \\
          \specialrule{0em}{3pt}{3pt}
          \hline
          \specialrule{0em}{1pt}{1pt}
          \textit{ROOM}          &10            & 6          &\textbf{4}                      &4\\
          \specialrule{0em}{1pt}{1pt}
          \textit{room0}  & 30         & 27           & \textbf{23}                    &23\\
         \specialrule{0em}{1pt}{1pt}
         \textit{office0}&18          & 15           & \textbf{16} &17                    \\
          \specialrule{0em}{1pt}{1pt}
          \bottomrule
        \end{tabular}
      \end{table}

\begin{figure}[!ht]
        \centering
        \includegraphics[width=3.5in]{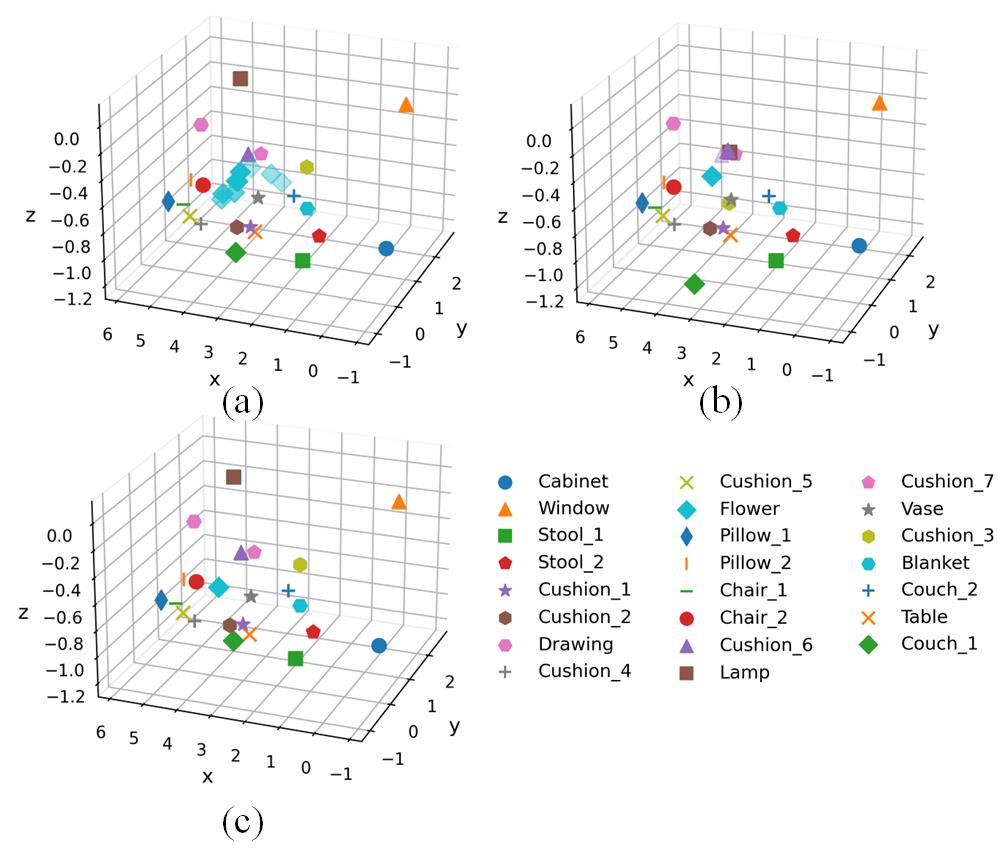}
        \caption{Qualitative comparison of data association results. (a) Only IoU
method. (b)Only QD method. (c)IOU conbined QD method. }
        \label{ass1}
 \end{figure}

In this section, we validate the effectiveness of the proposed data associdation strategy.
As Table \ref{ASSOCIATION} and Fig.\ref{ass1} show, the performance of different data association strategies, only IoU, only QD, and a QD+IoU, are systematically compared across three distinct scenes. The results indicate that the QD+IoU approach outperforms the individual metrics, demonstrating that integrating both metrics enhances the accuracy of association.

% \begin{figure}[!h]
%   \centering
%   \subfigure[]{
%     \includegraphics[width=4cm]{fig/IoU_QD_fig/only_IoU.png}
%   }
%   \subfigure[]{
%     \includegraphics[width=4cm]{fig/IoU_QD_fig/only_QD.png}
%   }
%   \subfigure[]{
%     \includegraphics[width=4cm]{fig/IoU_QD_fig/iou_QD.png}
%   }
%   \subfigure{
%     \includegraphics[width=4cm]{fig/IoU_QD_fig/legend_2.png}
%   }
%   \caption{Qualitative comparison of data association results. (a) Only IoU
% method. (b)Only QD method. (c)IOU conbined QD method. }
%   \label{ass1}
% \end{figure}

\section{CONCLUSIONS}
We present DQO-MAP, a novel object SLAM system tightly integrate object pose estimation and reconstruction. Our approach employs 3D Gaussian Splatting for object reconstruction and leverages quadrics for precise pose estimation. While Gaussian management and object association are handled on the CPU, all components are optimized in parallel on the GPU, significantly enhancing system efficiency. Comprehensive experiments demonstrate that our system excels in both object reconstruction and pose estimation.  In the future,  we plan to focus on leveraging  object maps for downstream tasks such as object navigation,  robotic manipulation and scene understanding.

%\addtolength{\textheight}{-12cm}   
%\addtolength{\textheight}{-12cm}


%%%%%%%%%%%%%%%%%%%%%%%%%%%%%%%%%%%%%%%%%%%%%%%%%%%%%%%%%%%%%%%%%%%%%%%%%%%%%%%%

%%%%%%%%%%%%%%%%%%%%%%%%%%%%%%%%%%%%%%%%%%%%%%%%%%%%%%%%%%%%%%%%%%%%%%%%%%%%%%%%



%\vfill
\bibliographystyle{unsrt}
\bibliography{IEEEexample}

\end{document}